%% file: ACVR2020_CameraReady.tex
\documentclass[runningheads]{llncs}
\usepackage{graphicx}

\usepackage{tikz}
\usepackage{comment}
\usepackage{amsmath,amssymb} 
\usepackage{color}
\usepackage{hyperref}

\begin{document}
\pagestyle{headings}
\mainmatter
\def\ECCVSubNumber{100}  

\title{Exploiting Scene-specific Features for Object Goal Navigation} 

\titlerunning{Exploiting Scene-specific Features for Object Goal Navigation}
%
\author{Tommaso Campari\inst{1,2}\orcidID{0000-0002-0435-4397} \and
Paolo Eccher\inst{1}\orcidID{0000-0002-5094-7767} \and
Luciano Serafini\inst{2}\orcidID{0000-0003-4812-1031} \and 
Lamberto Ballan\inst{1}\orcidID{0000-0003-0819-851X}}
\authorrunning{T. Campari et al.}
%
\institute{Department of Mathematics “Tullio Levi-Civita”, University of Padova, Italy \and
Fondazione Bruno Kessler, Trento, Italy\\
\email{tommaso.campari@phd.unipd.it}}
\maketitle

\begin{abstract}
Can the intrinsic relation between an object and the room in which it is usually located help agents in the Visual Navigation Task?
We study this question in the context of Object Navigation, a problem in which an agent has to reach an object of a specific class while moving in a complex domestic environment.
In this paper, we introduce a new reduced dataset that speeds up the training of navigation models, a notoriously complex task. Our proposed dataset permits the training of models that do not exploit online-built maps in reasonable times even without the use of huge computational resources. Therefore, this reduced dataset guarantees a significant benchmark and it can be used to identify promising models that could be then tried on bigger and more challenging datasets.
Subsequently, we propose the SMTSC model, an attention-based model capable of exploiting the correlation between scenes and objects contained in them, highlighting quantitatively how the idea is correct.
\end{abstract}

\input{Chapter/Intro.tex}

\input{Chapter/Relwork.tex}

\input{Chapter/Dataset.tex}

\input{Chapter/Model.tex}

\input{Chapter/Experiments.tex}

\input{Chapter/Results.tex}

\input{Chapter/Conclusion.tex}

\clearpage

%
%


\input{ACVR2020_CameraReady.bbl}
\end{document}

%% file: Chapter/Intro.tex
\section{Introduction}
Visual Navigation is a trending topic in the Computer Vision research community.
This growth in interest is undoubtedly due to the important practical implications that the development of agent capable of moving in complex environments can have on our society. For example, in an ever closer future we will be able to ask robotic assistants to perform the most disparate tasks in our homes.
Before we can ask a robot to take something out of the refrigerator, however, we need to make sure that it is able to find the refrigerator and get to it while avoiding the complex tangle of obstacles that a domestic environment can contain.
For this reason, this work focuses on the Object Navigation task, defined in \cite{anderson2018evaluation} as the search for objects belonging to a specific class by a robotic agent.
For humans, this is a very simple task whatever the object to be found is. A human can build a mental link between the object and the room where it is more likely to be found. In this way a human is able to simplify the problem by first searching for the room and then for the required object inside the room. For example, just think of having to look for a sink, unconsciously we will first search for the kitchen or bathroom, and then we will find a sink in them. To implement this intuition we have developed an attention-based \cite{vaswani2017attention} policy in which we exploit a joint-representation that integrates inside it visual informations extracted with a scene classifier and encoding of the semantic goal to be searched. This representation allows us to have a significant increase in performance compared to other models taken into consideration.\\
Several works tried to tackle Navigation through Learning models \cite{chaplot2020object}\cite{chang2020semantic}. In particular, \cite{chaplot2020object} leveraged depth images to construct in an online fashion semantic maps of the environment. From these maps they tried to maximize the exploration of the scene. To do this, they placed intermediate subgoals in unexplored areas of the map that the agent was encouraged to reach through planning. \\
This type of approach inevitably tends to lengthen the agent's paths, at least until the object sought is clearly visible, since wanting to maximize exploration involves a significant amount of moves by the agent.
For the simpler PointGoal Navigation \cite{anderson2018evaluation} task, proposed in the ``Habitat Challenge 2019''\footnote{\href{https://aihabitat.org/challenge/2019/}{https://aihabitat.org/challenge/2019/}}, it was possible to observe how a simple model \cite{wijmans2019dd} based on LSTM was able to perform better than competitors based on more complex architectures that exploit maps creation \cite{gupta2017cognitive}\cite{chaplot2020learning}.
This was made possible by the DD-PPO algorithm \cite{wijmans2019dd}, a distributed version of the PPO \cite{schulman2017proximal} Reinforcement Learning algorithm capable of parallelizing learning in a massive way.
In fact, they used 64 NVIDIA V100 GPUs for 3 consecutive days of training. In other words, the model was trained for about 180 days in a single GPU setting. Not all researchers, however, can have access to those massive hardware resources. For this reason, we have generated a reduced version of the dataset produced for the ``Habitat Challenge 2020''\footnote{\href{https://aihabitat.org/challenge/2020/}{https://aihabitat.org/challenge/2020/}} for the Object Navigation task that would allow the training of Deep Reinforcement Learning models in a few hours. In this way, even using few computational resources, complex models can be trained which on the original dataset would take days.\\
Furthermore, in \cite{fang2019scene} it was pointed out how in the Navigation tasks the use of Recurrent Neural Networks is not recommended. These structures usually have the issue of considering as more important the recent past and at the same time gradually forgetting the remote past.
On the contrary, by exploiting the principle of attention, described in \cite{vaswani2017attention}, it is possible to record all the past observations in a memory from which to extract information on every single step that the agent has undertaken in the past, improving that highly penalizing intrinsic aspect in the behavior of the Recurrent Neural Networks.\\
Summarizing, our contribution is twofold:
\begin{itemize}
    \item We propose a new reduced dataset for the Object Navigation task extracted from the one proposed in the ``Habitat Challenge 2020'', on which it is possible to test algorithms that would require a lot of resources for training and that maintain as far as possible the main characteristics that the first had;
    \item We propose the SMTSC model, an attention-based policy for Object Navigation that is able to exploit, starting from RGB images only, the idea mentioned above, namely that there exist a correlation that binds objects to specific rooms. This intuition improves performance, as demonstrated by the results obtained on a preliminary study performed using the aforementioned dataset.
\end{itemize}

%% file: Chapter/Relwork.tex
\section{Related Works}
Visual Navigation is an increasingly central topic within Computer Vision. However, Visual Navigation involves several different sub-problems and, in this section, we will summarize the most relevant related works to \textit{Simulators and 3D Datasets for Visual Navigation} and to other different research areas connected to \textit{Visual Navigation}.

\subsection{Simulators and 3D Datasets for Visual Navigation}
In recent years a large number of different simulators have been developed. GibsonEnv \cite{GIBSONENV}\cite{xia2020interactive} and AI2Thor \cite{kolve2017ai2} both allow to simulate multi-agent situations and to interact with objects, for example lifting them, pushing them, etc. Matterport3DSimulator \cite{anderson2018vision} can provide the agent with photorealistic images extracted from Matterport3d \cite{MATTERPORT3D} and is mainly used for Room2Room Navigation problem. HabitatAI \cite{savva2019habitat}, instead, provides support to 3D datasets such as Gibson, Matterport3D and Replica \cite{replica19arxiv}.
\subsection{Visual Navigation}
Also thanks to the new possibilities offered by these simulators today there are numerous tasks available, as pointed out in \cite{anderson2018evaluation}. Common Navigation tasks are mainly divided into two categories, namely those that require active exploration of the environment and those that, on the other hand, provide tools that can signal, for example via GPS sensors, the direction to be taken to reach the required goal.\\
In Classical Navigation, there are numerous approaches that perform path planning on explicit maps \cite{kavraki1996probabilistic}\cite{canny1988complexity}\cite{lavalle2001rapidly}.\\
More recently, however, approaches based on Reinforcement Learning have been presented through policies based on Recurrent Neural Networks \cite{mirowski2016learning}\cite{lample2017playing}\cite{savva2017minos}\cite{fang2019scene}\cite{mousavian2019visual}. Mirowski et al. \cite{mirowski2016learning} define an approach that jointly learns the goal-driven Reinforcement Learning problem with auxiliary depth prediction and loop closure classification tasks by exploiting the A3C algorithm \cite{mnih2016asynchronous}. Mousavian et al. \cite{mousavian2019visual} propose a Deep Reinforcement Learning framework that uses an LSTM-based policy for Semantic Target Driven Navigation. But LSTMs when they have to analyze very long data sequences tend to focus more on the most recent observations, giving less importance to the first ones that have been seen.
On the contrary, Fang et al. \cite{fang2019scene} propose Scene Memory Transformer, a policy based on attention \cite{vaswani2017attention} that is able to exploit even the least recent steps performed by the agent. In this case, the training of the policy is performed through the Deep Q-Learning algorithm \cite{mnih2015human}.
Starting from the work done in \cite{zamir2018taskonomy}, Sax et al. \cite{sax2019learning} show that using Mid-Level Vision results in policies that learn faster and generalize better when compared to learning from scratch. The Mid-Level model achieved high results in the PointGoal Navigation task. In \cite{wijmans2019dd}, a scalable Reinforcement Learning algorithm on multiple GPUs capable of solving the PointGoal Navigation task almost perfectly has been presented. This solution, in particular, shows how Visual Navigation is a really complex task that requires an impressive amount of resources. In fact, their training was conducted on 64 GPUs for 3 days. Unfortunately, these resources are not within the reach of the whole scientific community and training similar models remain almost prohibitive for most researchers.\\
For the Target Driven Navigation some works have recently been presented, such as \cite{wortsman2019learning}\cite{yang2018visual}\cite{wu2018learning}. Wu et al. \cite{wu2018learning} construct a probabilistic graphical model over the semantic information to explore structural similarities between the environment. Yang et al. \cite{yang2018visual}, instead, propose a Deep Reinforcement Learning model that exploits the relationships between objects, encoded through a Graph Convolutional Network, to incorporate semantic priors within the policy. Chaplot et al. \cite{chaplot2020object} propose a model for the ObjectGoal Navigation that constructs, during the exploration, a map with the semantic information of the scene extracted through a semantic segmentation model; from the generated map a long-term goal is selected to maximize exploration, through a policy trained with Reinforcement Learning, when the searched object is visible it is set as a new long-term goal. The actions of the agent are selected through the use of the FastMarching algorithm \cite{sethian1996fast}.

%% file: Chapter/Dataset.tex
\section{Dataset}
\label{Chap:Dataset}
\begin{figure}[t!]
    \centering
    \includegraphics[width=\textwidth]{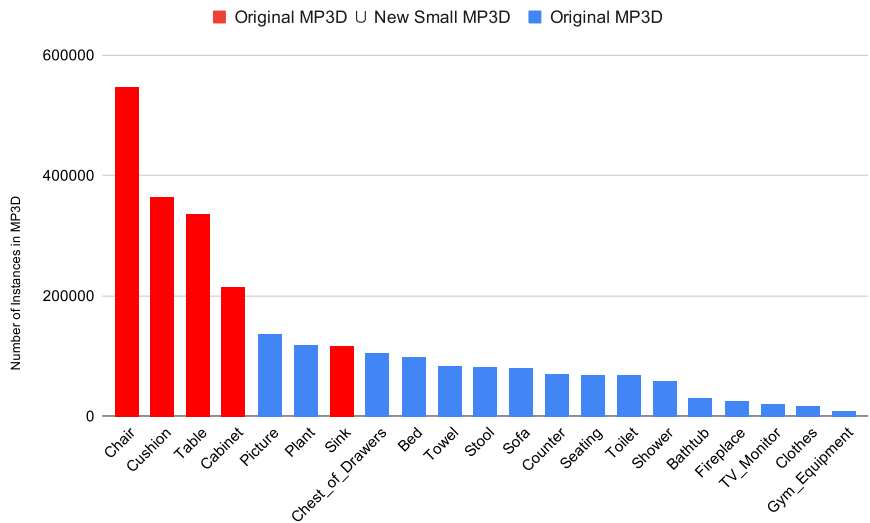}
    \caption{Distribution of objects in the Matterport3D Dataset.}
    \label{fig:OBJDISTRIBUTION}
\end{figure}

Previous works \cite{wijmans2019dd} have been able to achieve excellent results on the Point Navigation task \cite{anderson2018evaluation}, a simpler Navigation problem that doesn't require Semantic capabilities, while using an architecture that didn't include complex components such as occupancy maps \cite{chen2019learning}. However, they leveraged massive parallelism using hardware resources that are inaccessible to most institutions. 
We investigate on the possibility of solving the same problem in a reduced dataset with a minimal set of computational and time resources. For this reason we decided to concentrate on a subset of the Matterport3D Dataset \cite{MATTERPORT3D}. We argue that our choice of such subset still offers significant results as its statistical indicators are similar to the one of original Matterport3D. \\
The extraction of the subset was done restricting the problem to 5 out of 21 objects, choosing \texttt{Chair, Cushion, Table, Cabinet, Sink}. These objects are among the most frequent in the original set as is shown in Figure \ref{fig:OBJDISTRIBUTION}. We decided to include the \texttt{Sink} object as it is a characterising element of the bathroom. This increase the diversity of the domestic environments represented by our proposed Dataset. \\
Related to the distributions of objects, one of the main issue in the evaluation of Object Navigation agents using Matterport3D, is the Long Tail Distribution that limits the number of instances of infrequent objects that are seen during training. This, combined with metrics that ignore the precision on single classes, may undermine the development of agents with truly semantic capabilities.
Furthermore, we decided to extract our new dataset - from here we will refer to it as \textit{Small MP3D} - using 6 out of 56 scenes of the official Training split. These scenes are \texttt{r47D5H71a5s, i5noydFURQK, ZMojNkEp431, jh4fc5c5qoQ, HxpKQynjfin} and \texttt{GdvgFV5R1Z5}. For each object in each scene we extracted 100 episode for the training split and 20 for both validation and test. In total, Small MP3D possesses 3000 training episodes and 600 episodes for both validation and test. To ensure the ability to generalize to unseen scenes, we decided to generate also an Unseen Test and Validation Set, extracted from the \texttt{D7N2EKCX4Sj} and \texttt{aayBHfsNo7d} scenes, each with 100 episodes (10 episodes for each object class in each scene). \\
The characteristics of the new dataset are shown in Table \ref{tab:dataset1}. We report the average values of Euclidean and Geodesic Distance as well as the number of Steps required to complete the episodes. Overall, the complexity of the training split of the proposed reduced dataset is lesser than the original data as both Geodesic distance and the required Number of Steps are lower. However, the complexity of the Unseen Test split is significantly higher, with 30\% more required steps on average. We think that this additional complexity can guarantee a fair and meaningful benchmark.

\begin{table}[t!]
\centering
\begin{tabular}{l|r|r|r|r|r|r|r|r|r|}
\cline{2-10}
                                               & \multicolumn{3}{c|}{Chair}                                                       & \multicolumn{3}{c|}{Cushion}                                                     & \multicolumn{3}{c|}{Table}                                                                \\ \cline{2-10} 
                                               & \multicolumn{1}{c|}{Euc} & \multicolumn{1}{c|}{Geo} & \multicolumn{1}{c|}{Steps} & \multicolumn{1}{c|}{Euc} & \multicolumn{1}{c|}{Geo} & \multicolumn{1}{c|}{Steps} & \multicolumn{1}{c|}{Euc} & \multicolumn{1}{c|}{Geo} & \multicolumn{1}{c|}{Steps}          \\ \hline
\multicolumn{1}{|l|}{Original Train}           & 5.17                     & 6.45                     & 40.05                      & 5.39                     & 7.23                     & 45.56                      & 4.98                     & 6.19                     & 38.82                               \\ \hline
\multicolumn{1}{|l|}{\textbf{New Train}}       & \textbf{3.50}            & \textbf{4.00}            & \textbf{27.71}             & \textbf{5.79}            & \textbf{7.23}            & \textbf{49.14}             & \textbf{3.12}            & \textbf{3.85}            & \textbf{28.42}                      \\ \hline
\multicolumn{1}{|l|}{Original Val}             & 5.09                     & 7.21                     & 43.32                      & 4.50                     & 6.18                     & 38.77                      & 3.66                     & 5.29                     & 33.83                               \\ \hline
\multicolumn{1}{|l|}{\textbf{New Seen Val}}    & \textbf{3.43}            & \textbf{3.89}            & \textbf{27.01}             & \textbf{6.09}            & \textbf{7.58}            & \textbf{51.37}             & \textbf{2.77}            & \textbf{3.42}            & \textbf{26.42}                      \\ \hline
\multicolumn{1}{|l|}{\textbf{New Unseen Val}}  & \textbf{4.07}            & \textbf{5.79}            & \textbf{37.15}             & \textbf{9.00}            & \textbf{12.19}           & \textbf{68.40}             & \textbf{4.53}            & \textbf{5.18}            & \textbf{32.95}                      \\ \hline
\multicolumn{1}{|l|}{\textbf{New Seen Test}}   & \textbf{3.64}            & \textbf{4.15}            & \textbf{28.76}             & \textbf{6.25}            & \textbf{7.61}            & \textbf{50.17}             & \textbf{3.15}            & \textbf{3.79}            & \textbf{27.82} \\ \hline
\multicolumn{1}{|l|}{\textbf{New Unseen Test}} & \textbf{4.06}            & \textbf{5.11}            & \textbf{32.75}             & \textbf{10.13}           & \textbf{12.94}           & \textbf{70.50}             & \textbf{4.55}            & \textbf{5.37}            & \textbf{34.45}                      \\ \hline
                                               & \multicolumn{3}{c|}{Cabinet}                                                     & \multicolumn{3}{c|}{Sink}                                                        & \multicolumn{3}{c|}{Total Average}                                                        \\ \cline{2-10} 
                                               & \multicolumn{1}{c|}{Euc} & \multicolumn{1}{c|}{Geo} & \multicolumn{1}{c|}{Steps} & \multicolumn{1}{c|}{Euc} & \multicolumn{1}{c|}{Geo} & \multicolumn{1}{c|}{Steps} & \multicolumn{1}{c|}{Euc} & \multicolumn{1}{c|}{Geo} & \multicolumn{1}{c|}{Steps}          \\ \hline
\multicolumn{1}{|l|}{Original Train}           & 5.24                     & 6.84                     & 42.63                      & 6.33                     & 8.54                     & 51.80                      & 5.27                     & 6.78                     & 42.27                               \\ \hline
\multicolumn{1}{|l|}{\textbf{New Train}}       & \textbf{4.02}            & \textbf{4.91}            & \textbf{34.15}             & \textbf{4.94}            & \textbf{5.99}            & \textbf{39.80}             & \textbf{4.27}            & \textbf{5.19}            & \textbf{35.84}                      \\ \hline
\multicolumn{1}{|l|}{Original Val}             & 5.40                     & 7.46                     & 46.10                      & 6.52                     & 8.83                     & 53.50                      & 4.73                     & 6.65                     & 40.98                               \\ \hline
\multicolumn{1}{|l|}{\textbf{New Seen Val}}    & \textbf{4.02}            & \textbf{4.87}            & \textbf{34.07}             & \textbf{4.62}            & \textbf{5.69}            & \textbf{38.22}             & \textbf{4.19}            & \textbf{5.09}            & \textbf{35.42}                      \\ \hline
\multicolumn{1}{|l|}{\textbf{New Unseen Val}}  & \textbf{5.80}            & \textbf{6.76}            & \textbf{46.65}             & \textbf{7.17}            & \textbf{8.87}            & \textbf{52.25}             & \textbf{6.11}            & \textbf{7.76}            & \textbf{47.48}                      \\ \hline
\multicolumn{1}{|l|}{\textbf{New Seen Test}}   & \textbf{4.37}            & \textbf{5.33}            & \textbf{36.58}             & \textbf{4.96}            & \textbf{6.06}            & \textbf{40.16}             & \textbf{4.47}            & \textbf{5.39}            & \textbf{36.70}                      \\ \hline
\multicolumn{1}{|l|}{\textbf{New Unseen Test}} & \textbf{8.40}            & \textbf{9.58}            & \textbf{56.85}             & \textbf{8.87}            & \textbf{10.93}           & \textbf{63.60}             & \textbf{7.20}            & \textbf{8.79}            & \textbf{51.63}                      \\ \hline
\end{tabular}
\label{tab:dataset1}
\caption{Statistics of our dataset}
\end{table}

%% file: Chapter/Model.tex
\section{Method}
\label{Chapter:Method}
In this section we first describe the Problem Setup. Then we introduce our SMTSC model as shown in Fig. \ref{fig:MODEL}
\subsection{Problem Setup}
Our interests fall within the task of Object Navigation. In particular, this task requires finding an occurrence of a certain class starting from a random position in the environment.\\
This task can be viewed as a Partially Observable Markov Decision Process (POMDP)\cite{kaelbling1998planning}\cite{fang2019scene} $(\mathcal{S}, \mathcal{A},\mathcal{O}, R(s,a), T(s'|s,a), P(o|s))$ in which:
\begin{itemize}
    \item $\mathcal{S}$ is a finite set of states of the world;
    \item $\mathcal{A}$ is a finite set of actions;
    \item $\mathcal{O}$ is the observation spaces;
    \item $R: \mathcal{S} \times \mathcal{A} \rightarrow \mathbb{R}$ it is a reward function, which given a state \textit{s} and an action \textit{a} to be performed in it, returns a reward for the execution of \textit{a} in \textit{s}. 
    \item $T: \mathcal{S} \times \mathcal{A} \rightarrow \mathcal{S}$ it is a transition function, which given a state \textit{s} and an action \textit{a} to be performed in it, returns the probability of reaching the state \textit{s}' by executing \textit{a} in \textit{s};
    \item $P(o|s)$ is a probability density function that defines the likelihood of observing \textit{o} in \textit{s}.
\end{itemize}
In the setup taken into consideration the set of possible actions is defined as $\mathcal{A} = \{$ \textbf{go\_forward, turn\_left, turn\_right, stop} $\}$. 
The actions are deterministic, that is, apart from possible collisions with objects in the scene, the agent will move in the desired direction without deviations due to noisy dynamics. In particular, with a go\_forward action the agent will move forward by $0.25m$, while with the two turn actions, it will rotate $30^{\circ}$ in the desired direction. Finally, a stop-action causes the navigation episode to end and, if the agent is less than $0.1m$ from an object of the type sought, then the episode will be deemed successfully concluded. \\
The observation $o = ($RGB, p, a$_{\text{prev}}$, goal$) \in \mathcal{O}$ is the set of features collected by the agent at each step in the environment and passed to the model. RGB is what the agent sees from a given position, it is an RGB image extracted with 640x480 size; \textit{p} is the agent's position w.r.t. to the starting point, $a_{\text{prev}}$ is the action performed in the previous step and finally \textit{goal} is the objective object to be sought.
\subsection{Model}
\begin{figure}[t!]
    \centering
    \includegraphics[width=\textwidth]{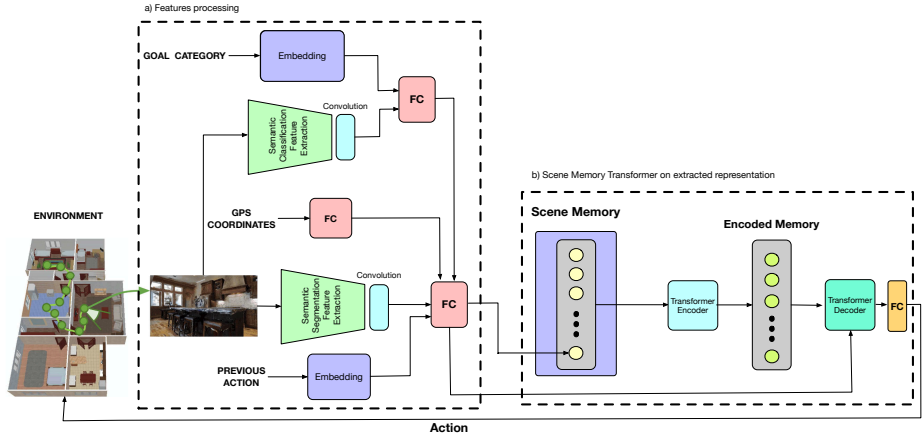}
    \caption{SMTSC model: a) Features processing: all features are processed and a shared representation is created. This representation is then inserted into a memory (b) Scene Memory Transformer) from which the action that the agent will perform is extracted with a Scene Memory Transformer model.}
    \label{fig:MODEL}
\end{figure}
The proposed model is visible in Figure \ref{fig:MODEL}, it is composed of two main parts, a first part in which the features are extracted and brought into a joint representation and a second module in which the features of the current observation are added to a memory that keeps track of all past observations and an attention-based policy network extracts a distribution on possible actions.
\subsubsection{Features Processing Module}
Starting from an observation $o = ($RGB, p, a$_{\text{prev}}$, goal$) \in \mathcal{O}$ we first define 5 different encoders:
\begin{enumerate}
    \item $\gamma_{\text{sem\_seg}}: \mathbb{R}^{256\times256} \rightarrow \mathbb{R}^{256}$: encodes an RGB observations into a vector of size 256 by using features extracted from a semantic segmentation model.
    \item $\gamma_{\text{scene\_class}}: \mathbb{R}^{256\times256} \rightarrow \mathbb{R}^{128}$: encodes an RGB observations into a vector of size 128 by using features extracted from a scene classification model.
    \item $\gamma_{\text{goal}}(goal)$: encodes the goal into a vector of size 32
    \item $\gamma_{\text{pos}}(p)$: encodes the relative position into a vector of size 32
    \item $\gamma_{\text{act}}(a_{\text{prev}})$: encodes the previous action executed into a vector of size 32
\end{enumerate}
Now we define:
\begin{equation}
    \delta(RGB, goal)=FC(\{\gamma_{\text{scene\_class}}(RGB),\gamma_{\text{goal}}(goal)\})
    \label{eq:1}
\end{equation}
and:
\begin{equation}
    \phi(o)=FC(\{\gamma_{\text{sem\_seg}}(RGB), \gamma_{\text{pos}}(p), \gamma_{\text{act}}(a_{\text{prev}}), \delta(RGB, goal)\})
    \label{eq:2}
\end{equation}
Eq. \ref{eq:1} encodes the previously illustrated idea that a goal is intrinsically associated with a specific room. To do this starting from the goal through the $\gamma_{\text{goal}}$ function, a representation of dimension 32 is extracted. In parallel, a scene classifier is used to extract features starting from the RGB image, these two modalities are concatenated and a joint representation is created using a fully connected layer. In this way, we obtain a representation of the goal conditioned step by step from the room in which the agent is located, that is going to add useful information to the agent to understand how to move in the environment.\\
Finally, Eq. \ref{eq:2} generates a joint representation between all the modalities described above. It, therefore, concatenates the representation obtained from a model for semantic segmentation and the representations of the previous action, position and goal (as defined by Eq. \ref{eq:1}). This vector is passed to a fully connected layer which returns a vector of size 256.
\subsubsection{Scene Memory Transformer Module}
\begin{figure}[t!]
    \centering
    \includegraphics[width=\textwidth]{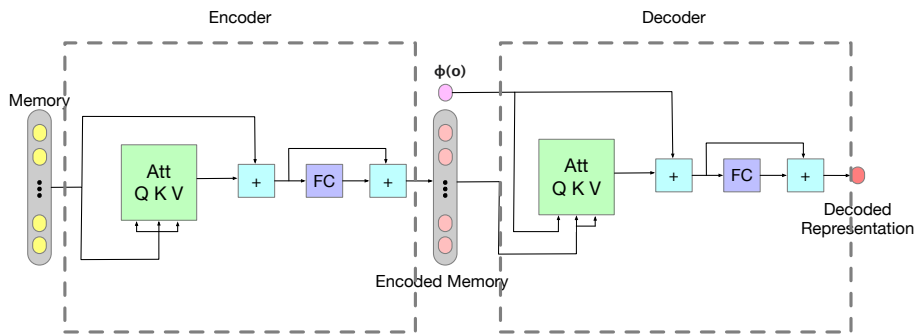}
    \caption{The structure of the encoder-decoder model based on Multi-Head Attention, as presented in Fang et al.\cite{fang2019scene}.}
    \label{fig:encdec}
\end{figure}
SMT module is based on the one proposed by \cite{fang2019scene}. Given a new episode, we build an initially empty memory where to save the joint-representations obtained starting from observation $o=(\text{RGB, p, a}_{\text{prev}}\text{, goal })$ through the application of the $\phi(o)$ function defined in the Eq. \ref{eq:2}. This memory is then passed along with $\phi(o)$ to an attention-based encoder-decoder policy \cite{vaswani2017attention} which extracts a probability distribution on the actions. The encoder-decoder structure is shown in Figure \ref{fig:encdec}. The SMT encoder uses a self-attention to encode the M memory, so M is passed to a MultiHeadAttention with 8 heads in which therefore M is both Query and Key and Value as shown in Eq. \ref{eq:3}.
\begin{equation}
    M' = \text{MultiHeadAttention}(M, M)
    \label{eq:3}
\end{equation}
Subsequently, M is passed together with the joint-representation to the decoding structure. It always uses the attention mechanism to give a representation of $\phi(o)$ conditioned by past observations. Again, attention is implemented through an 8-head MultiHeadAttention mechanism as shown in Eq. \ref{eq:4}.
\begin{equation}
    Q = \text{MultiHeadAttention}(\phi (o), M')
    \label{eq:4}
\end{equation}
Finally, \textit{Q} is reduced to the dimensionality of the action space $\mathcal{A}$ through a Linear Layer and a Categorical Distribution on the action space is extracted from the latter representation, as defined in Eq \ref{eq:5}.
\begin{equation}
    \pi(a|o, M)=\text{Categorical}(\text{Softmax}(\text{FC}(Q)))
    \label{eq:5}
\end{equation}
\subsubsection{Implementation and Training Details}
We implemented all the models using Python 3.6 and PyTorch \cite{paszke2017automatic}. We used the PPO \cite{schulman2017proximal} algorithm to train the model using a 32 GB Tesla V100 GPU. We used batch size of 64 and Adam Optimizer with a Learning Rate of $1 \times 10^{-5}$. The visual features are extracted from the images using the Taskonomy networks \cite{zamir2018taskonomy} as done in \cite{sax2019learning}. This allows us to have consistent features as Taskonomy networks have been trained on indoor environments simulated by Gibson \cite{GIBSONENV} and also the number of parameters to be trained on the network drops drastically. All the activation functions within the encoder-decoder structure are ReLU functions and the pose vector is encoded as a quadruple $(x, y, sin(\theta), cos(\theta))$.

%% file: Chapter/Experiments.tex
\section{Experimental Results}
In this section an accurate description of the experimental setup will be provided, the results obtained will then be presented.
\subsection{Experimental setup}
We used the \textit{Small MP3D} dataset presented in Section \ref{Chap:Dataset}.
We decided to test the SMTSC model on both the seen and unseen test sets. Results on the seen set will give as a measure of its capacity to memorize environments seen during training. Conversely, results obtained on the unseen set will quantify the degree of adaptation to unseen scenes that the agent possess. However, as the number of training scenes is only six, we don't expect our agents to develop strong generalization behaviours.
The simulator used for all the experiments was Habitat, which provides the agent with 640x480 RGB images. In addition to the images, an odometry system is available that can provide the x and y coordinates and the orientation of the agent with respect to the starting point. The simulated robot has a height of 88 cm from the ground with a radius of 18cm. The camera of the agent with which he acquires the images is placed 88cm from the ground and allows to capture images with a 79$^{\circ}$ HFOV. Sliding against objects is not allowed, once a collision has been made the agent must necessarily rotate before being able to proceed again in the environment. The moves that the agent can perform at each step are: 25cm move forward, 30$^{\circ}$ turn left, 30$^{\circ}$ turn right and stop. In particular, when the stop action is called the current episode is declared correct only if the object sought is less than 0.1m from the agent. For each episode the agent has a maximum of 500 steps to call the stop action, otherwise the episode is considered to be a failure automatically.\\
The metrics used to evaluate the proposed models are 3: Success Rate, Success weighted by Path Length (SPL) and distance to success (DTS).
The Success Rate is simply the ratio between the number of episodes that have been successful and the total number of episodes. The SPL, on the other hand, measures the efficiency in reaching the goal when an episode is successfully completed with a numerical value between 0 and 1. When the episode is not successfully completed 0 is attributed to this metric, otherwise it can be calculated by using Eq. \ref{eq:6}, in which \textit{N} is the number of test episodes, \textit{l$_i$} is the shortest-path distance from the agent’s starting position to the goal in episode \textit{i}, \textit{p$_i$} the length of the path actually taken by the agent in the episode \textit{i} and finally \textit{S$_i$} is a binary indicator of success.
\begin{equation}
    \text{SPL}=\frac{1}{N}\sum_{i=1}^NS_i\frac{\text{\textit{l}}_i}{\text{max}(p_i,\text{\textit{l}}_i)}
    \label{eq:6}
\end{equation}
The SPL is considered today as the main metric for the Object Navigation task, but as described in \cite{batra2020objectnav} there are numerous problems. In fact, not all the failures are to be considered equal, just think of how an agent arrived at 0.2m from the object sought is evaluated with the same score obtained by another agent who has rotated on himself for the entire duration of the episode. Finally, DTS is defined as the agent's distance from the threshold boundary around the nearest object. This is mathematically defined as:
\begin{equation}
    DTS = max(\text{Geo(AgentPos, GoalPos)}-d, 0)
\end{equation}
In which Geo function measures the geodesic distance between the agent at the end of the episode and the nearest object, \textit{d} instead is the success distance, in this case 0.1m.\\
Apart from the model presented in Section \ref{Chapter:Method}, four other different baselines have been tested to assess the performance of the proposed model.\\
\textbf{Random Agent:} an agent that performs random actions extracted from a uniform distribution.\\
\textbf{Forward-Only Agent:} an agent who only performs forward actions (with a 1\% probability of calling the stop action). These first two baselines were placed to demonstrate the non-triviality of the dataset proposed.\\
\textbf{Reactive Agent}: a policy that extracts semantic segmentation features through Taskonomy and merges them with position and goal as was done in \cite{sax2019learning}. The action to be performed is directly extracted from this representation, therefore no type of memory is used.\\
\textbf{LSTM Agent}: the representation is extracted as for the reactive agent, but in this case, it is passed to an LSTM and the action to be performed is extracted from its output. The model is pretty similar to the RGB-Only presented in \cite{wijmans2019dd}.\\
\textbf{SMT without Scene Classification Features (SMT w/o SC)}: this is the same model presented in Section \ref{Chapter:Method}, except for the fact that the goal is coded only through an embedding layer without exploiting the joint representation with the features of the Scene Classifier. \\
The last three models were trained with PPO Reinforcement Learning algorithm. 

%% file: Chapter/Results.tex
\subsection{Results}
\begin{table}[t!]
    \centering
    \begin{tabular}{|c|c|c|c|}
         \hline
         Model & SPL$\uparrow$ & Success$\uparrow$ & DTS$\downarrow$
         \\\hline
         Random       & 0.00 & 0.00 & 5.126
         \\\hline
         Forward-Only & 0.009 & 0.026 & 5.094
         \\\hline
         Reactive \cite{sax2019learning} & 0.041 & 0.126 & 4.523
         \\\hline
         LSTM \cite{wijmans2019dd} & 0.131 & 0.247 & 3.199
         \\\hline
         SMT w/o SC (our) & 0.345 & 0.595 & 1.562 
         \\\hline
         SMTSC (our) & \textbf{0.649} & \textbf{0.883} & \textbf{0.403}
         \\\hline
    \end{tabular}
    \caption{Results on the Seen Test Set}
    \label{tab:seen}
\end{table}

\begin{table}[t!]
    \centering
    \begin{tabular}{|c|c|c|c|}
         \hline
         Model & SPL$\uparrow$ & Success$\uparrow$ & DTS$\downarrow$
         \\\hline
         Random       & 0.00 & 0.00 & 7.842
         \\\hline
         Forward-Only & 0.004 & 0.01 & 7.922
         \\\hline
         Reactive \cite{sax2019learning} & 0.001 & 0.04 & 7.797
         \\\hline
         LSTM \cite{wijmans2019dd} & 0.002 & 0.04 & 7.648
         \\\hline
         SMT w/o SC (our) & 0.008 & 0.04 & 7.518 
         \\\hline
         SMTSC (our) & \textbf{0.039} & \textbf{0.080} & \textbf{6.817} 
         \\\hline
    \end{tabular}
    \caption{Results on the Unseen Test Set}
    \label{tab:unseen}
\end{table}

In Table \ref{tab:seen} are shown the results obtained on the test set in seen environments. The low performance of Random and Forward-Only baselines highlight how the dataset is not trivial, showing that the object to be found is almost never in the immediate proximity of the starting point of the episode. Looking instead at the other baselines based on Reinforcement Learning, we can see how the two models that use the Scene Memory Transformer are able to perform much better in almost all metrics. This big difference is probably attributable to the fact that the SMT can extract crucial information even from actions performed in a fairly remote past, while for example in the case of the LSTM this is very difficult, as more recent information tends to supplant the older ones and this is emphasized especially in very long sequences.\\
It is also interesting to note that the model that creates a joint representation between the goal coding and the visual features for the scene classification performs much better than the SMT w/o SC model. Even in the SPL, we have an increase of 88\%, in the success ratio an increase of 48.4\% and finally, the average DTS has fallen by over a meter.\\
The behavior observed in the seen test set follows the same trend also in the case of the unseen test set whose results are visible in Table \ref{tab:unseen}. In fact, taking as a comparison the average geodesic distance of the unseen test set (7.76m), we can see how the Forward Only and Random agents tend to conclude their episodes even further away from the starting point. The reactive model instead certifies its performance in line with the average distance reported in Table \ref{tab:dataset1} for the unseen test set. Finally, the two models that exploit the SMT are the ones that perform better, even in an unseen environment. In particular, the proposed model capable of exploiting the joint representation between visual features and goal coding lowers the average distance from the goal by almost a meter, and certifies its successes on 8\% of the scenes.\\
In the next section we report some qualitative example of the SMTSC model on the seen and unseen test sets.

\subsubsection{Qualitative Results}
\begin{figure}[t!]
    \centering
    \includegraphics[width=\textwidth]{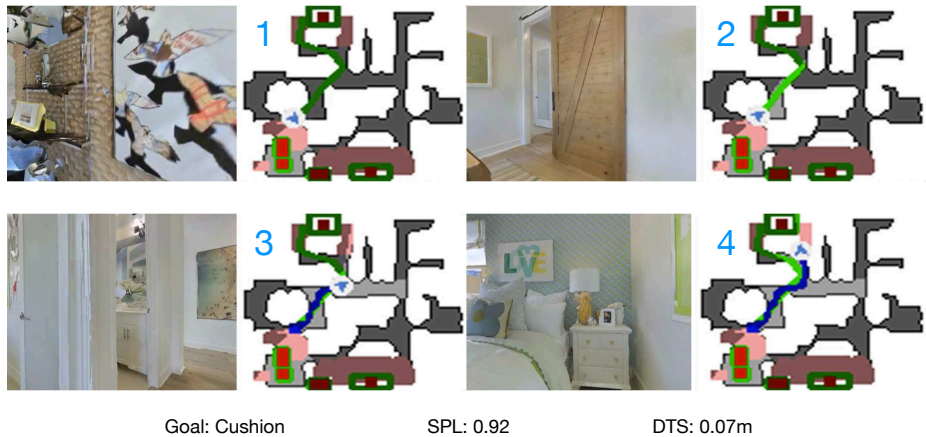}
    \caption{Successful navigation episode with SMTSC model on seen test set.}
    \label{fig:seentrue}
\end{figure}

\begin{figure}[t!]
    \centering
    \includegraphics[width=\textwidth]{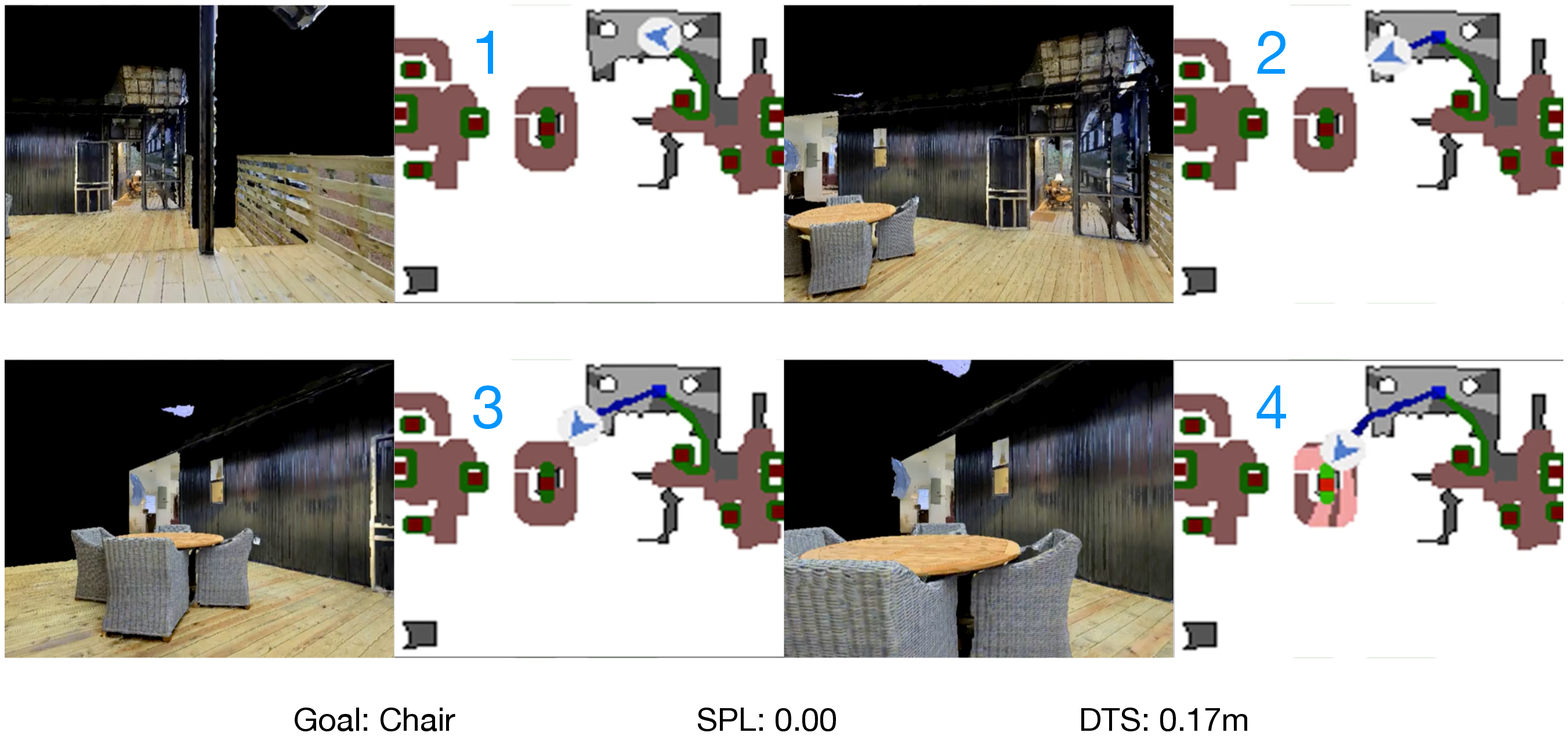}
    \caption{Unsuccessful navigation episode with SMTSC model on seen test set.}
    \label{fig:seenfalse}
\end{figure}

\begin{figure}[t!]
    \centering
    \includegraphics[width=\textwidth]{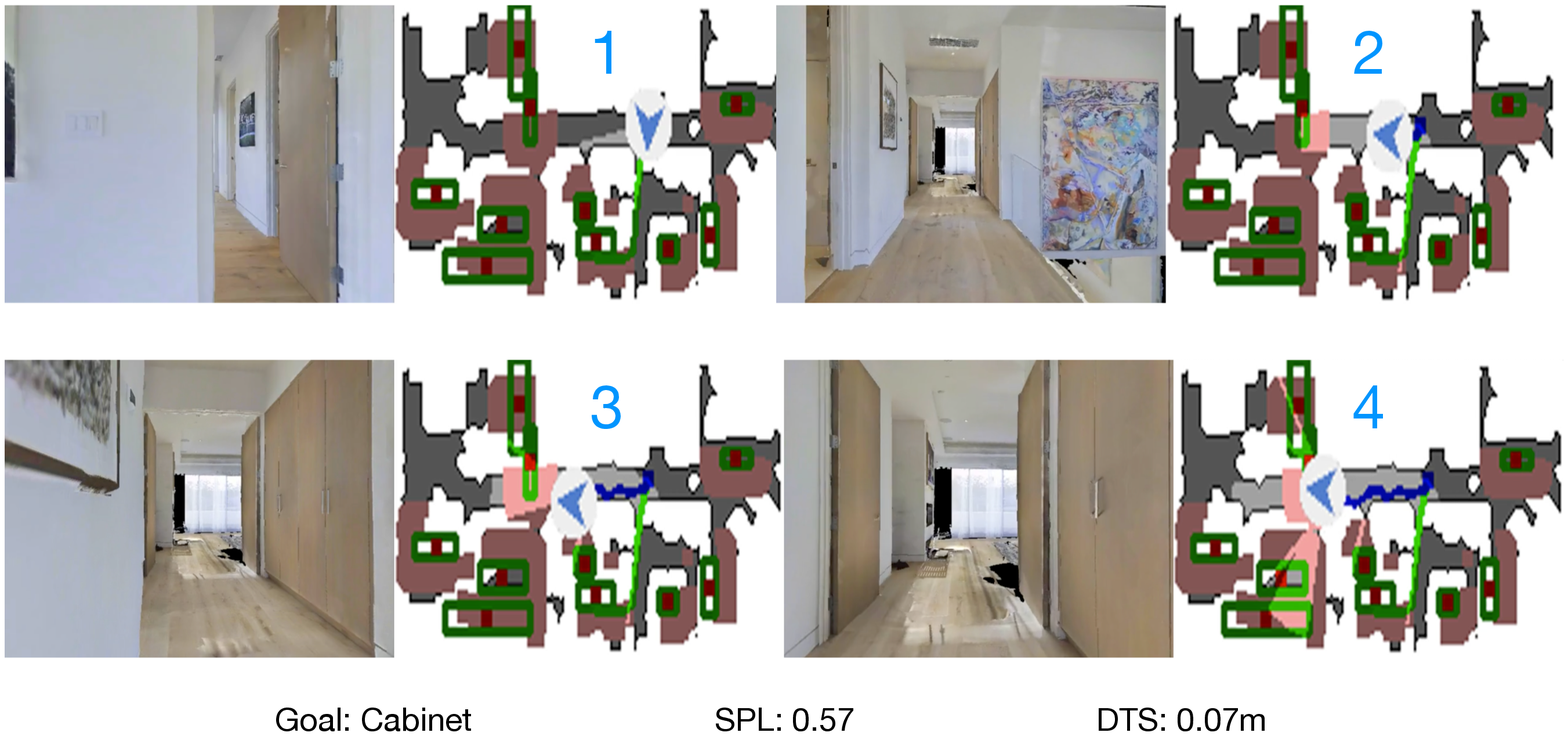}
    \caption{Successful navigation episode with SMTSC model on unseen test set.}
    \label{fig:unseentrue}
\end{figure}

\begin{figure}[t!]
    \centering
    \includegraphics[width=\textwidth]{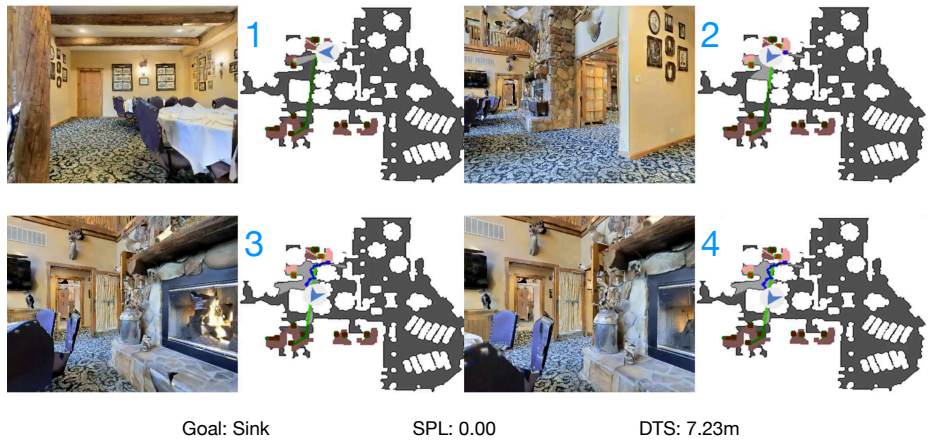}
    \caption{Unsuccessful navigation episode with SMTSC model on unseen test set.}
    \label{fig:unseenfalse}
\end{figure}

In Figure \ref{fig:seentrue} it is possible to see, in blue, the path taken by the agent to reach an object of the ``cushion'' class on the seen test set. In green you can see the shortest path to the object. The agent has successfully reached the object sought by stopping less than 0.1m from it. On the contrary, in Figure \ref{fig:seenfalse} is shown an example of failure, still on the seen test set, in which the agent was unable to reach the object sought. The agent, from his initial position was able to recognize an object of the class sought and to head towards it, despite not being the closest chair. At the time of calling the stop action, however, the agent was 7cm beyond the boundary that would have given the episode success. In this case, we can see the strong penalty given by a metric like the SPL, in fact, in our opinion, this episode cannot be considered totally wrong.\\
In Figures \ref{fig:unseentrue} and \ref{fig:unseenfalse}, on the other hand, it is possible to see two examples extracted from the evaluation of the model on the unseen test set. In the first image, the was able to take a correct path to a ``cabinet''. However, the cabinet that was found wasn't the nearest so the SPL of this episode was only 0.57. In the second example, on the other hand, a very long route of over 14 meters of navigation is presented. The agent here was able to follow the optimal path for about half of its length, then it reached the maximum number of actions allowed. This means that it ''wasted'' a lot of action to increase its understanding of the scene. 


%% file: Chapter/Conclusion.tex
\section{Conclusion}
We proposed a subset of the dataset developed for the Habitat 2020 ObjectNav Challenge \cite{savva2019habitat}. This was done to allow the training of models that do not involve the use of planning associated with the construction of maps within them with few computational resources (e.g. a single GPU). Furthermore, we proposed a model capable of exploiting the subtle relationship existing between objects and the rooms in which they are usually located. This intuition, combined with the use of a Scene Memory Transformer showed good results on the proposed dataset. In the future, it would be very interesting to be able to test this model on the complete dataset using the distributed Reinforcement Learning algorithm DD-PPO \cite{wijmans2019dd} and a greater number of GPUs in order to perform training in a reasonable time.